# Toward a Computational Theory of Evidence-Based Reasoning for Instructable Cognitive Agents

**Gheorghe Tecuci, Dorin Marcu, Mihai Boicu, Steven Meckl, Chirag Uttamsingh**

Learning Agents Center, Volgenau School of Engineering, George Mason University, Fairfax, VA 22030, USA
{ tecuci; dmarcu; mboicu; smeckl; cuttamsi }@gmu.edu

**Abstract**

Evidence-based reasoning is at the core of many problem-solving and decision-making tasks in a wide variety of domains. Generalizing from the research and development of cognitive agents in several such domains, this paper presents progress toward a computational theory for the development of instructable cognitive agents for evidence-based reasoning tasks. The paper also illustrates the application of this theory to the development of four prototype cognitive agents in domains that are critical to the government and the public sector. Two agents function as cognitive assistants, one in intelligence analysis, and the other in science education. The other two agents operate autonomously, one in cybersecurity and the other in intelligence, surveillance, and reconnaissance. The paper concludes with the directions of future research on the proposed computational theory.

## Introduction

Evidence is any observable sign, datum, or item of information that is relevant in deciding whether a statement or hypothesis under consideration is true or false.

English philosopher Jeremy Bentham, over two centuries ago, once famously said, "The field of evidence is no other than the field of knowledge" (Bentham, 1810). Indeed, evidence-based reasoning (EBR) is at the core of many problem-solving and decision-making tasks in a wide variety of domains, including cybersecurity, intelligence analysis, forensics, medicine, law, history, archaeology, and all the natural sciences (physics, chemistry, biology, etc.).

However, evidence-based reasoning has been studied in each of these domains in isolation from the others. Only recently steps have been taken to study evidence as a multidisciplinary subject (Twinning, 2003), and to lay the foundation for a *Science of Evidence* (Schum, 2009).

Generalizing from our research and development of cognitive agents in several domains, we are making progress toward formulating a computational theory supporting the development of instructable cognitive agents for evidence-based reasoning tasks. Currently this theory includes:

- A systematic approach to evidence-based reasoning consisting of collaborative computational processes of evidence in search of hypotheses or questions in search of answers, hypotheses in search of evidence, and evidentiary testing of hypotheses.
- A general ontology of evidence and associated computational methods for assessing the credibility of different types of evidence.
- The architecture for an instructable cognitive agent shell that incorporates this general evidence-based reasoning knowledge, can be taught reasoning strategies, and can function either interactively, as an assistant to a human user, or autonomously.

This paper presents these aspects of the theory and their application to the development of four prototype cognitive agents in domains that are critical to the government and the public sector. Two of the agents function as cognitive assistants, one in intelligence analysis, and the other in science education. The other two agents operate autonomously, one in cybersecurity and the other in intelligence, surveillance, and reconnaissance.

## Evidence-Based Reasoning as Discovery of Evidence, Hypotheses, and Arguments

Developed in the framework of the scientific method, the computational approach to evidence-based reasoning views this process as *ceaseless discovery of evidence, hypotheses, and arguments* in a non-stationary world, involving collaborative computational processes of *evidence in search of hypotheses*, *hypotheses in search of evidence*, and *evidentiary testing of hypotheses* (see Figure 1).

First, through *abductive (imaginative) reasoning* that shows that something is *possibly* true, one generates alternative hypotheses that may explain an observation of interest or answer an important question. Next, through *deductive reasoning* that shows that something is *necessarily* true, one uses these hypotheses to generate new lines of inquiry and discover new evidence for testing them. After that, through *inductive reasoning* that shows that something is *probably* true, one tests the hypotheses by developing argumentation structures that show how the discovered evidence favors or disfavors them.

*Evidence in search of hypotheses*, *hypotheses in search of evidence*, and *evidentiary testing of hypotheses* are collaborative processes that support each other in recursive calls, as shown in the bottom part of Figure 1. For example, the discovery of new evidence may lead to the modification of the existing hypotheses or the generation of new ones that, in turn, lead to the search and discovery of new evidence. Also, inconclusive testing of the hypotheses leads to the need of discovering additional evidence.

In the following we discuss several general methods for performing each of these processes, either by a person in collaboration with a cognitive assistant, or by an autonomous agent.

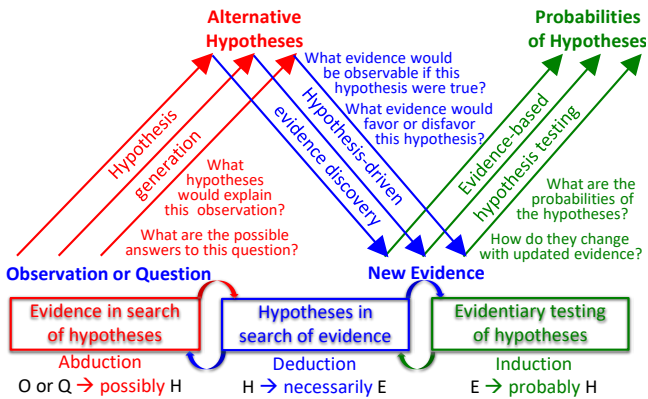

*Figure 1: The evidence-based reasoning processes.*

## Evidence in Search of Hypotheses

The EBR process is usually triggered by an interesting observation that needs to be explained. The question is: *What hypotheses would explain this observation?* Answering this question is challenging both for humans and for automated agents because it involves abductive (imaginative) reasoning. Automatic hypothesis generation through abductive reasoning is computationally-intensive because there are numerous hypotheses that can be abduced from an observation. Eco (1983) distinguishes between four types of abduction based on their level of creativity (see the left-hand side of Table 1). In *overcoded abduction* we have evidence that event E occurred and prior knowledge "H→E" from which we infer that H is possible. In *undercoded abduction* our prior knowledge includes multiple rules ($H_1$→E, $H_2$→E, … $H_n$→E) and we decide to select the antecedent of one as our hypothesis. In *creative abduction* we have no relevant prior knowledge but we have a "hunch" that H might be true. Finally, in *meta-abduction*, we have a sequence of abduction steps, from E to F, from F to G, and from G to H, and we decide to act on H without having a chance to verify each stage of reasoning.

Thagard (1993) identified four other types of abduction, based on what is being abduced (see the top part of Table 1). In *simple abduction* we try to explain why some entity, object, or person X has an interesting characteristic or behavior G, and we make the hypothesis that X is an H because all entities in H have G. In *existential abduction*, we hypothesize the existence of some previously unobserved entity which is in a certain relation R with the entity having the characteristic or behavior G to be explained, because the existence of R explains G. In *analogical abduction*, we have already generated some initial hypothesis H. But in the past when H was true, K was also true. Therefore, we can refine our hypothesis to H & K. Finally, in *rule-forming abduction*, we hypothesize a rule that allows us to explain why the entity X has that interesting characteristic or behavior G.

Schum, (2001a) showed that the types of abductions identified by Eco and Thagard are, in fact, orthogonal since we can identify all the possible 16 combinations of abductions, based on their creativity and form (i.e., *simple overcoded abduction* through *rule-forming meta abduction*), as shown in Table 1. While some of these abductions require human imaginative reasoning (e.g., *simple creative abduction*), those combining Thagard's *simple, existential, and analogical abduction*, with Eco's *overcoded and undercoded abduction* can be automatically performed (e.g., *analogical undercoded abduction*).

We are investigating a more efficient approach to hypothesis generation as a multi-step abductive process where each abductive step involving the generation of competing hypotheses is followed by evidence collection and testing of these hypotheses, to significantly prune the hypothesis space. The envisioned approach is illustrated in Figure 2. If we were to perform a single-step abduction, from evidence E to a hypothesis that would explain it, we would obtain a huge number of hypotheses represented as dots at the top of the figure. We would then need to investigate each of these competing hypotheses to find the most likely explanation.

Now consider performing multi-step abduction. From E one can abduce F, Fi, and Fj. At this point, we would search for evidence relevant to these three hypotheses and we would test them based on the discovered evidence, as illustrated in Figure 1, concluding, for example, that F is the most promising. Then we would continue the abduction from F, abducing G, Gm, and Gn, testing these hypotheses, and concluding, for example, that G is the most promising. Finally, from G, we would abduce H, Hp, and Hg, and test them. Such an approach based on spiral hybrid reasoning, where small abductive, deductive, and inductive steps feed off of each other, has the potential of significantly pruning the hypothesis space to tractable levels.

*Table 1: Species of Abduction.*

|  | simple | existential | analogical | rule-forming |
|---|---|---|---|---|
| overcoded | simple overcoded | existential overcoded | analogical overcoded | rule-forming overcoded |
| undercoded | simple undercoded | existential undercoded | analogical undercoded | rule-forming undercoded |
| creative | simple creative | existential creative | analogical creative | rule-forming creative |
| meta | simple meta | existential meta | analogical meta | rule-forming meta |

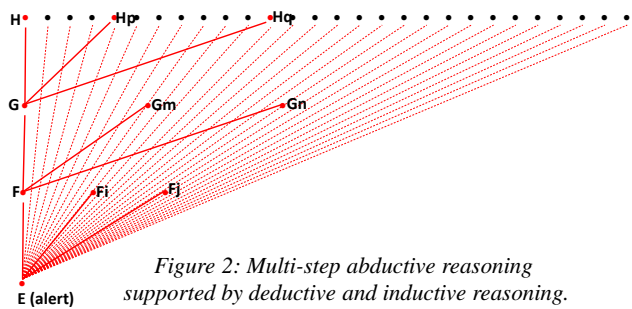

*Figure 2: Multi-step abductive reasoning supported by deductive and inductive reasoning.*

## Hypotheses in Search of Evidence

To discover new evidence, the agent puts each of the generated hypothesis to work, guiding the collection of relevant evidence. One approach is to ask the question: *What evidence would be observable if this hypothesis were true?* One decomposes the hypothesis into simpler and simpler hypotheses, and uses the simplest hypotheses to generate new lines of inquiry and discover new evidence. The reasoning might go as follows:

If $H_k$ were true then the sub-hypotheses $H_{k1}$, $H_{k2}$, and $H_{k3}$ would also need to be true. But if $H_{k2}$ were true then one would need to observe evidence $E_{k2}$, and so on (see the middle side of Figure 1 and Figure 3). This process leads to the discovery of new evidence by identifying the necessary conditions for hypothesis $H_k$.

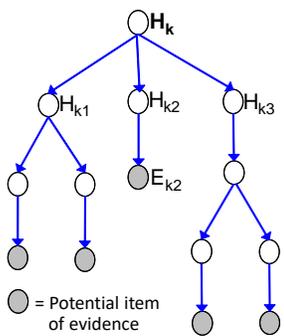

*Figure 3: Hypothesis in search of evidence.*

A broader question that guides the discovery of evidence is: *What evidence would favor or disfavor this hypothesis?* In this case one would look for *sufficient conditions*, or even *indicators*, for a hypothesis to be true or false. As we discuss next, this hypothesis decomposition process leads to the development of a Wigmorean probabilistic inference network for testing the hypothesis.

## Evidentiary Testing of Hypotheses

Hypothesis testing is probabilistic in nature because the evidence is always *incomplete* no matter how much we have and is commonly *inconclusive* in the sense that it is consistent with the truth of more than one hypothesis. Further, the evidence is frequently *ambiguous*, with multiple meanings. A mass of evidence is in most situations *dissonant,* some favoring and some disfavoring the hypothesis under consideration. Finally, the evidence comes from sources with different levels of *credibility*.

However, none of the non-enumerative probability views known to us (Subjective Bayes, Belief Functions, Fuzzy, and Baconian) can optimally cope with all these five characteristics of evidence (Schum, 2001b; Tecuci et al., 2016a). For example, both the Subjective Bayesian view and the Baconian view have difficulties with ambiguities or imprecision in evidence. On the other hand, the Fuzzy view and the Belief Functions view can naturally cope with such imprecisions. The Baconian view is the only one that can optimally deal with the incompleteness of evidence. The Subjective Bayesian, Belief Functions, and Fuzzy views all answer the question: *How strong is the evidence we do have about this hypothesis?* It is thus possible to encounter situations where, based on the current evidence, all these three views predict that $H_3$ is the most likely hypothesis, just to learn later that $H_1$ was the one that was true. The Baconian view would have helped with this situation because it answers the question: *How much evidence do we have about this hypothesis, and how many questions about it remain unanswered?* Thus, it would have helped to acknowledge that the answers to these unanswered questions may not favor $H_3$. Therefore, when testing a hypothesis, one should assess not only the probability of being true, but also the confidence in this probability.

The conclusion of the above discussion is that one should employ hypothesis testing methods that integrate different probability views, to take advantage of their complementary capabilities and cope with all the five characteristics of evidence. We have developed such a method that integrates logic and Baconian probabilities with Fuzzy qualifiers, and uses the min/max probability combination rules common to the Baconian and Fuzzy views of probability. That is, the probability of a conjunction of hypotheses is the minimum of their probabilities, and the probability of a disjunction of hypotheses is the maximum of their probabilities. In this method, hypotheses are tested by developing Wigmorean probabilistic inference networks like the one in Figure 4.

Hypothesis H is decomposed into simpler hypotheses by considering both favoring arguments (supporting the truth of H), under the left (green) square, and disfavoring arguments (supporting the falsehood of H), under the right (pink) square. Each argument is an independent strategy of showing that H is true or false, and is characterized by a specific relevance or strength. The argument consists either of a single hypothesis (e.g., H3) or a conjunction of hypotheses (e.g., H1 & H2). The hypotheses from these arguments are further decomposed through other arguments, leading to simpler and simpler hypotheses that can be more accurately assessed based on evidence.

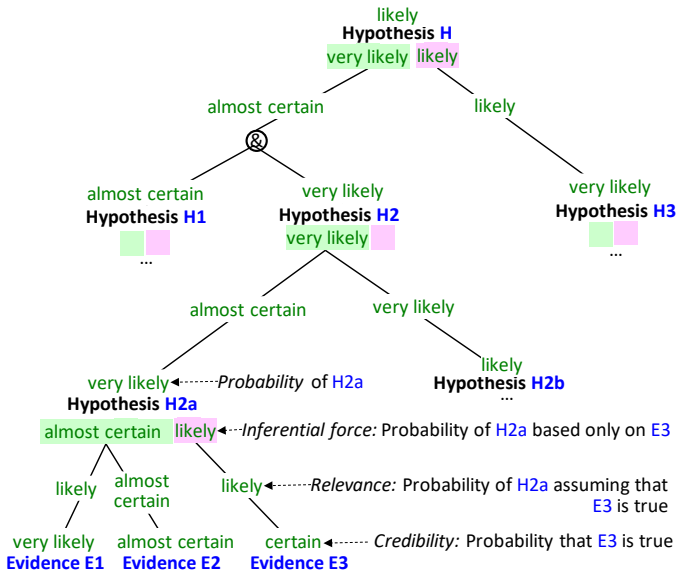

*Figure 4: Wigmorean probabilistic inference network.*

Consider, for example, hypothesis H2a from the bottom of Figure 4. It has two items of favoring evidence, E1 and E2, and one item of disfavoring evidence, E3. For each item, one assesses its credibility and its relevance to the hypothesis H2a. Then the inferential force of each item of evidence is computed as the minimum between its credibility and its relevance. The probability of hypothesis H2a is determined by balancing the combined inferential force of the favoring evidence (which is the maximum between the inferential force of E1 and that of E2), and the inferential force of the disfavoring evidence E3.

The probabilities of the upper-level hypotheses are computed in a similar way, based on the probabilities of their sub-hypotheses, leading to an assessment of the top level hypothesis H.

## General EBR Knowledge

An evidence-based reasoning agent needs a significant amount of knowledge in order to perform the processes illustrated in Figure 1. Part of it, however, is general evidence-based reasoning knowledge that is defined once and can be used by any agent, as discussed below.

## Ontology of Evidence

Figure 5 shows a fragment of a general ontology of evidence (Schum et al., 2009). *Tangible evidence* is evidence that can be directly examined to see what events it might reveal. *Real tangible evidence* is the thing itself, while *demonstrative tangible evidence* is a representation or illustration of a thing (e.g., a diagram, map, sensor image, or record of some sort). *Testimonial evidence* is evidence about some event provided by a human source. It can be based on direct observation, second hand report from another source, or on the basis of opinion or inferences based on information about the occurrence of other events. An *authoritative record* is evidence accepted to be true, such as a tide table, celestial table, or table of physical or mathematical results. Finally, *missing evidence* is evidence that is expected to exist but is not found.

## Credibility Assessment Patterns

For each evidence type in the ontology from Figure 5 there is a general, domain-independent pattern for assessing its credibility. For illustration, let us consider a person S reporting the observation of an event that supports the truthfulness of hypothesis H. The evidence E provided by S is an example of *unequivocal testimonial evidence based upon direct observation* (see bottom-left of Figure 5). The credibility of E depends on the credibility of the source S that, in turn, depends on the source's *competence*, *veracity*, and *accuracy*, as shown in the top part of Figure 6. Each of these credibility indicators depends on lower level indicators. For example, the veracity of the source depends on the *truthfulness of the information* provided by the source and on the *trustworthiness of the source*. The truthfulness of the information provided by the source depends on *corroborative evidence* (Is there any other evidence that corroborates or confirms this source's report?) and on *contradictory evidence* (Is there any existing evidence that contradicts or conflicts with what the source has reported?).

The trustworthiness of the source depends on source's *character* (What evidence do we have about this source's character and honesty?), *reliability* (What does the record show about the truthfulness of this source's previous reports?), *goals* (Does what this source tells us support any of his or her goals?), and other indicators, as shown at the bottom right of Figure 6 (Tecuci et al., 2016a).

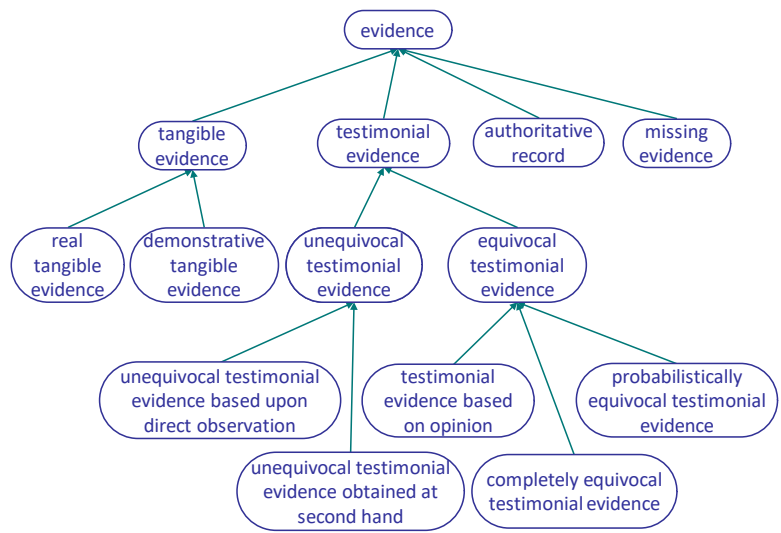

*Figure 5: Fragment of the ontology of evidence.*

The probabilities of the lowest-level indicators in Figure 6 are assessed based on evidence. Then these probabilities are composed, from bottom up, to obtain the probabilities of the upper level indicators and, ultimately, the probability that evidence E is true (i.e., the credibility of E).

In a given situation, however, we may not have evidence to evaluate each indicator. The pattern in Figure 6 shows, for each indicator, the relevance of each possible combination of the present sub-indicators. For example, the relevance of a source's competence, veracity, and accuracy taken together, on the source's credibility is certain (C). However, the relevance of source's competence and veracity is only likely (L), and the relevance of veracity alone is barely likely (BL). This explains the contradictions in the evidence on the same event, provided by the two main witnesses in the Senate confirmation hearings of Judge Kavanagh for the Supreme Court, where each of these witnesses were assessed as having high veracity. As one can see, when assessing the credibility of testimonial evidence, there is much more to consider than veracity. The * operator in Figure 6, called *combined indicator*, computes the probability of a hypothesis based on the presence and probability of its indicators, and corresponds to the disjunction of all possible conjunctions of these indicators.

## Instructable Cognitive Agent Shell

The computational theory of evidence-based reasoning described in the previous section has led to the design and development of the general instructable agent shell shown in Figure 7. The shell employs a hybrid knowledge representation consisting of an ontology and various types of rules, including the general knowledge discussed in the previous section. The shell has two main components, a *Mixed-Initiative Learning and Reasoning Assistant* and an *Autonomous Multi-Agent Reasoner*.

A subject matter expert directly teaches the Mixed-Initiative Learning and Reasoning Assistant how to investigate a specific alert by following the systematic process illustrated in Figure 1. During this process the expert explains each step involved in the investigation to the agent, from generating the hypotheses that may explain that alert, to using each of these hypotheses in guiding the collection of additional evidence, and to testing the hypotheses based on the collected evidence. Through the employment of the Disciple-EBR multistrategy learning approach, which integrates learning from examples, learning from explanations, and learning by analogy and experimentation, in a mixed-initiative interaction with the expert, the assistant learns general rules to generate hypotheses, rules to search for and monitor evidence, and rules to test the hypotheses. Successive versions of this learning approach are presented in (Tecuci 1998; Tecuci et al. 2005; 2008; 2016b; Boicu et al., 2011).

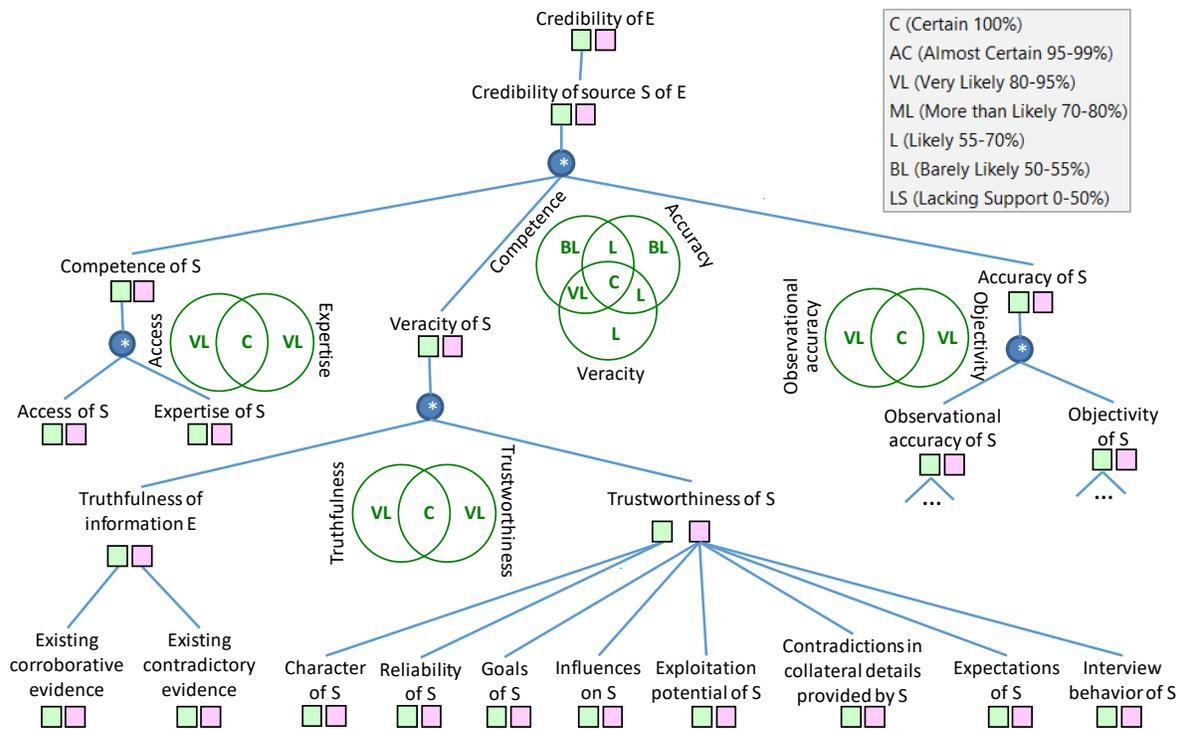

*Figure 6: Credibility pattern for a source of testimonial evidence based upon direct observation.*

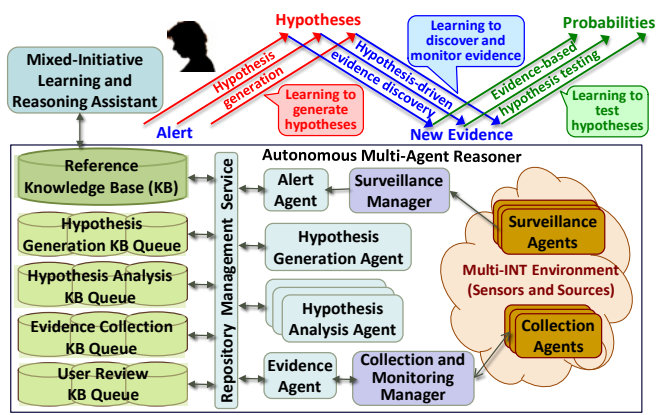

*Figure 7: Architecture of the learning agent shell.*

The result of teaching the agent is a Reference Knowledge Base that enables the Autonomous Multi-Agent Reasoner to perform evidence-based reasoning. The agents of the Autonomous Multi-Agent Reasoner are copies of the corresponding modules of the Mixed-Initiative Learning and Reasoning Assistant except that they are configured to run autonomously and communicate by developing and exchanging the Knowledge Bases shown in the left hand side of the Figure 7. Notice that the architecture may include multiple copies of an agent (e.g., the Hypothesis Analysis Agent) to speed-up the analysis process. Connection to the application environment is done through the Surveillance Manager and the Collection and Monitoring Manager.

## Evidence-Based Reasoning Agents

The presented computational theory and agent shell architecture emerged from our experience in developing several agents for evidence-based reasoning. In the following we overview four of these agents developed for domains that are critical to the government and the public sector, highlighting the specific challenges addressed by each. Features of these agents that are of particular interest to the government are: *transparency, explanations* and, *trust*, by making explicit their reasoning as Wigmorean probabilistic inference networks with symbolic probabilities that are easy to understand, and by their ability to self-analyze their reasoning, highlighting potential biases and weaknesses; *control*, through a mixed-initiative interaction allowing the user to decide the agent's level of autonomy, from fully automatic, to human "on the loop," to human "in the loop," and to complete user control; and *adaptability*, through continuous learning and refinement from domain experts, without requiring software development.

### Cogent: Cognitive Assistant for Intelligence Analysts

Intelligence analysts face the astonishingly complex task of drawing defensible and persuasive conclusions from masses of evidence of all types, in a world that is changing all the time. A prediction just made may be overtaken by subsequent events. Such predictions are often of immediate interest, without time for extensive research and deliberation on the available evidence which, moreover, is always incomplete, commonly inconclusive, frequently ambiguous, dissonant in most situations, and from sources with various levels of credibility. These evidence characteristics make the predictions, not only probabilistic in nature, but also prone to errors, because none of the probability systems (e.g., Subjective Bayesian, Belief Functions, Baconian, Fuzzy) can individually cope with all these five characteristics of evidence. Additional errors are introduced by the analyst's biases.

To address these challenges, we developed Cogent as an assistant for intelligence analysts, with the goal of enabling a synergistic integration of the analyst's imagination and expertise with the computer's knowledge and critical reasoning (Tecuci et al., 2015; 2018). Consider again the process described in Figure 1. While the analyst has to imagine the possible answers to the intelligence question asked, Cogent helps with producing a schematic diagram that completely lays out the underlying analytic framework for every analytic conclusion, including the connection between the evidence and various intermediate conclusions in the analysis, the evaluation of the credibility of evidence and its strength in supporting a conclusion, and the role of any assumptions in addressing missing information. Cogent can also detect several biases, such as *confirmation bias* (building an argumentation and only search for evidence that confirms analyst's beliefs while dismissing or ignoring evidence to the contrary), *satisficing bias* (choosing the first hypothesis that appears good enough, rather than carefully identifying all possible hypotheses and determining which one is the most consistent with the evidence), and potential *absence of evidence bias* (failure to consider the degree of completeness of the available evidence). Many other biases are avoided because explicit argumentations are developed and Cogent employs an intuitive system of symbolic probabilities. Additionally, Cogent facilitates the analysis of what-if scenarios, where the analyst may make various assumptions and Cogent automatically determines their influence on the analytic conclusion. It also automatically updates the analysis based on new or revised evidence. Once the analysis is finalized, Cogent generates a structured report that the analyst then transforms into a more understandable and persuasive production report. The final report, which includes argumentation fragments and evidence, can be shared with other users, subjected to critical analysis, and correspondingly improved.

Cogent and associated training lectures are accessible at http://lac.gmu.edu/cogent/

## sInvestigator: Science Investigator

Inquiry-based teaching in schools at all levels is recognized as being effective, but difficult to use in practice. The effort required to design inquiry-based lessons is greater than for content-based lessons because the teacher needs to both introduce the topics and to guide their discovery by the students. Even when inquiry-based lessons are already available, their adaptation to the knowledge and skill level of the students is challenging. Also challenging is the evaluation of the quality and effectiveness of inquiry because direct student comparisons are not usually possible.

sInvestigator is a cognitive assistant that was developed to help science teachers in conducting inquiry-based classes that help students develop critical thinking skills in addressing scientific problems. It has built-in features to engage the students in understanding, extending, creating, critiquing, and debating evidence-based scientific argumentations in real-life scientific investigations. For example, the instructor formulates an inquiry and the students imagine its possible answers. Then they search evidence on the Internet and develop argumentations for the competing answers to determine the most likely one.

We have experimentally used sInvestigator in HNRS 353 Modern Scientific Revolutions, HNRS 240 History of Science, and HNRS 353 Science of Cities, all George Mason

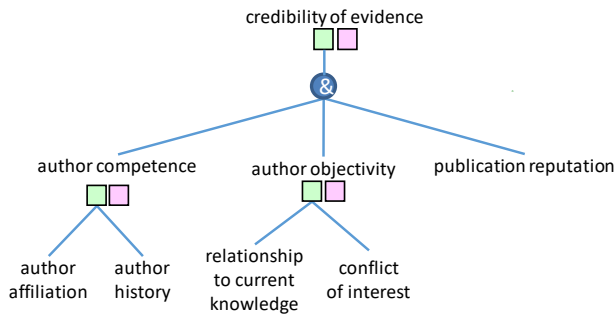

*Figure 8: Credibility assessment pattern in sInvestigator.*

University undergraduate science classes for non-science majors. Examples of inquiries investigated in these classes include: Has life changed over time? What is the mechanism by which life changes over time? How will fully autonomous vehicles make road transportation safer? What energy source is appropriate for San Francisco?

sInvestigator was particularly helpful in assisting students in performing deep credibility analyses of scientific evidence obtained from the Internet, by employing the pattern shown in Figure 8 which is a simplification and adaptation of the general credibility pattern from Figure 6. The agent is accessible at http://lac.gmu.edu/sInvestigator/

## CAPIP: Cognitive Agent for Persistent Intelligence Processing

Hypothesis generation and analysis in intelligence, surveillance, and reconnaissance (ISR) is not currently automated, which is a significant limitation for the ISR systems that continuously produce huge amounts of information in need of immediate processing. CAPIP is an initial prototype system designed for automatic ISR (Tecuci et al., 2019). It watches for alerts among the events provided by the MITRE's Integrated Environment for Persistent Intelligence software. An alert is an indicator of one or several hypotheses of interest, such as the notification that tracking of Ship1 by the Automatic Identification System was lost at Time1 near Location1. Possible explanatory hypotheses include: "Ship1 performs covert goods transfer", "Ship1 performs illegal fishing operations", and "Ship1 avoids tracking by pirates." CAPIP follows the EBR process from Figure 1 to generate alternative explanations of the alert and determine the most likely hypothesis.

## CAAPT: Cognitive Agents for Advanced Persistent Threats

Modern cyber defense is done in a cybersecurity operations center (CSOC) where analysts monitor alerts and log data from available information sources and use them to make determinations about the presence or absence of intrusion activity (Zimmerman, 2014). However, the large and increasing number of alerts and the time required for their manual analysis creates a very complex and expensive security environment for network defense organizations that are faced with a shortage of cybersecurity analysts and an ever increasing analyst cost. Among the biggest security challenges are those from the Advanced Persistent Threats (APTs), computer network exploitation groups (many of them state-sponsored) that leverage superior resources, knowledge, and tactics to gain and maintain access to targeted networks and adapt to defenders' efforts to resist them. Current automatic security systems do not reliably detect APTs because this requires reasoning over a large set of weak indicators. Therefore, the detection is manually done by experienced cyber analysts.

In an attempt to address these problems, we have developed CAAPT with the architecture shown in Figure 7 (Meckl et al., 2017). The Mixed-Initiative Learning and Reasoning Assistant is instructed by a cybersecurity expert how to investigate cybersecurity alerts. The resulting Reference Knowledge Base enables the Autonomous Multi-Agent System to investigate alerts, as the cyber expert would, but in a transparent manner that allows a natural "on the loop" supervision by a cyber operator. At the same time, CAAPT can also operate interactively, with the "user in the loop," as a trusted collaborator of the human analyst.

We have designed and performed experiments to test both the training of CAAPT and its ability to detect configuration changes in the same malware and new malware versions as the attackers' tool set evolved over time. The experiment simulated a subset of the historical evolution of APT1 malware: Auriga → Auriga variants → Bangat → Bangat variants → Seasalt → Seasalt variants → Kurton → Kurton variants (Mandiant, 2013). Attack models were tested in isolated virtual machines by simulating APT1 malware infections and running CAAPT against them. After being trained to detect a specific malware (e.g., Auriga), the agent was able to detect variants of that malware, and it required limited incremental training to detect other members of the family of that malware. Additionally, CAAPT rapidly detected APT1 intrusions through a rigorous and transparent analysis, as judged by the training expert.

## Conclusions

Generalizing from the research and development of instructable cognitive agents for evidence-based reasoning in several domains, this paper presented progress toward a computational theory that will improve and accelerate the development of such agents. It presented a framework for evidence-based reasoning, a domain-independent ontology of evidence, a sample credibility analysis pattern, and the architecture of an instructable agent shell. Future research will extend and improve this computational theory and the component methods for evidence-based reasoning. It will also develop the agent shell and the presented agents, and investigate new agents in domains such as personal health monitoring and fraud detection in the financial services.

## Acknowledgements

David Schum has significantly influenced this work that was supported in part by the National Science Foundation under grant number 1611742, by the Air Force Research Laboratory under contract number FA8750-17-C-0002, by MITRE Corporation under agreement number 114615, by IARPA under contract number 2017-16112300009, and by George Mason University. The views and conclusions contained herein are those of the authors and should not be interpreted as necessarily representing the official policies, either expressed or implied, of any organization of the U.S. Government. The U.S. Government is authorized to reproduce and distribute reprints for governmental purposes notwithstanding any copyright annotation therein.